\documentclass[letterpaper, 10 pt, conference]{ieeeconf}

\IEEEoverridecommandlockouts 

\usepackage[utf8]{inputenc}
\usepackage{graphicx} 
\usepackage{caption}
\usepackage{subcaption}

\DeclareCaptionFormat{myformat}{\fontsize{8}{8}\selectfont#1#2#3}
\usepackage{cite}
\usepackage{amsmath,amssymb,amsfonts}
\usepackage{algorithmic}
\usepackage[ruled,vlined]{algorithm2e}
\SetKwInOut{Input}{input}
\SetKwInOut{Output}{output}
\SetKwInOut{Parameter}{parameters}

\usepackage{float}
\usepackage{textcomp}
\PassOptionsToPackage{hyphens}{url}
\usepackage{dblfloatfix}
\usepackage{listings,multicol}

\title{\LARGE \bf
Grapevine Winter Pruning Automation: On Potential Pruning Points Detection through 2D Plant Modeling using Grapevine Segmentation
}

\author{Miguel Fernandes$^{1,2}$, Antonello Scaldaferri$^{1}$, Giuseppe Fiameni$^{3}$, Tao Teng$^{1,4}$, Matteo Gatti$^{4}$, Stefano Poni$^{4}$,\\ Claudio Semini$^{5}$,~\IEEEmembership{Member,~IEEE}, Darwin Caldwell$^{1}$,~\IEEEmembership{Senior Member,~IEEE}, Fei Chen$^{6}$,~\IEEEmembership{Senior Member,~IEEE}
\thanks{This research is supported by the project ``Grapevine Recognition, Manipulation and Winter Pruning Automation'' funded by IIT-Unicatt Joint Lab. \textit{(Corresponding author: Fei Chen)}}
\thanks{$^{1}$Miguel Fernandes, Antonello Scaldaferri, Tao Teng, Darwin Caldwell are with Active Perception and Robot Interactive Learning Laboratory, Department of Advanced Robotics, Istituto Italiano di Tecnologia, Via Morego 30, 16163, Genova, Italy (e-mail: {\tt\small name.surname@iit.it}).}%
\thanks{$^{2}$Miguel Fernandes is with Department of Informatics, Bioengineering, Robotics and System Engineering, Università di Genoa, Viale Causa 13, 16145 Genova, Italy (e-mail: {\tt\small miguel.ferreira@iit.it}).}%
\thanks{$^{3}$Giuseppe Fiameni is with NVIDIA AI Technology Center (NVAITC), Italy (email: {\tt\small gfiameni@nvidia.com}).}
\thanks{$^{4}$Tao teng, Matteo Gatti, Stefano Poni are with Department of Sustainable Crop Production, Università Cattolica del Sacro Cuore, Via Emilia Parmense 84, 29122 Piacenza, Italy (e-mail: {\tt\small name.surname@unicatt.it}).}%
\thanks{$^{5}$Claudio Semini is with Dynamic Legged Systems (DLS) lab, Istituto Italiano di Tecnologia, Via Morego 30, 16163, Genova, Italy (e-mail: {\tt\small name.surname@iit.it}).}%
\thanks{$^{6}$Fei Chen is with Department of Mechanical and Automation Engineering, T-Stone Robotics Institute, The Chinese University of Hong Kong, Chung Chi Rd, Ma Liu Shui, Hong Kong (e-mail: {\tt\small f.chen@ieee.org}).}%
}

\begin{document}
\setlength{\textfloatsep}{0.1cm}
\setlength{\floatsep}{0.1cm}
\maketitle
\thispagestyle{empty}
\pagestyle{empty}

\begin{abstract}
Grapevine winter pruning is a complex task, that requires skilled workers to execute it correctly.
The complexity of this task is also the reason why it is time consuming. Considering that this operation takes about 80-120 hours/ha to be completed, and therefore is even more crucial in large-size vineyards, an automated system can help to speed up the process.
To this end, this paper presents a novel multidisciplinary approach that tackles this challenging task by performing object segmentation on grapevine images, used to create a representative model of the grapevine plants. Second, a set of potential pruning points is generated from this plant representation.
We will describe (a) a methodology for data acquisition and annotation, (b) a neural network fine-tuning for grapevine segmentation, (c) an image processing based method for creating the representative model of grapevines, starting from the inferred segmentation and (d) potential pruning points detection and localization, based on the plant model which is a simplification of the grapevine structure.
With this approach, we are able to identify a significant set of potential pruning points on the canes, that can be used, with further selection, to derive the final set of the real pruning points.
\end{abstract}

\section{INTRODUCTION}

\begin{figure}[t]
    \centering
    \includegraphics[width=0.45\textwidth]{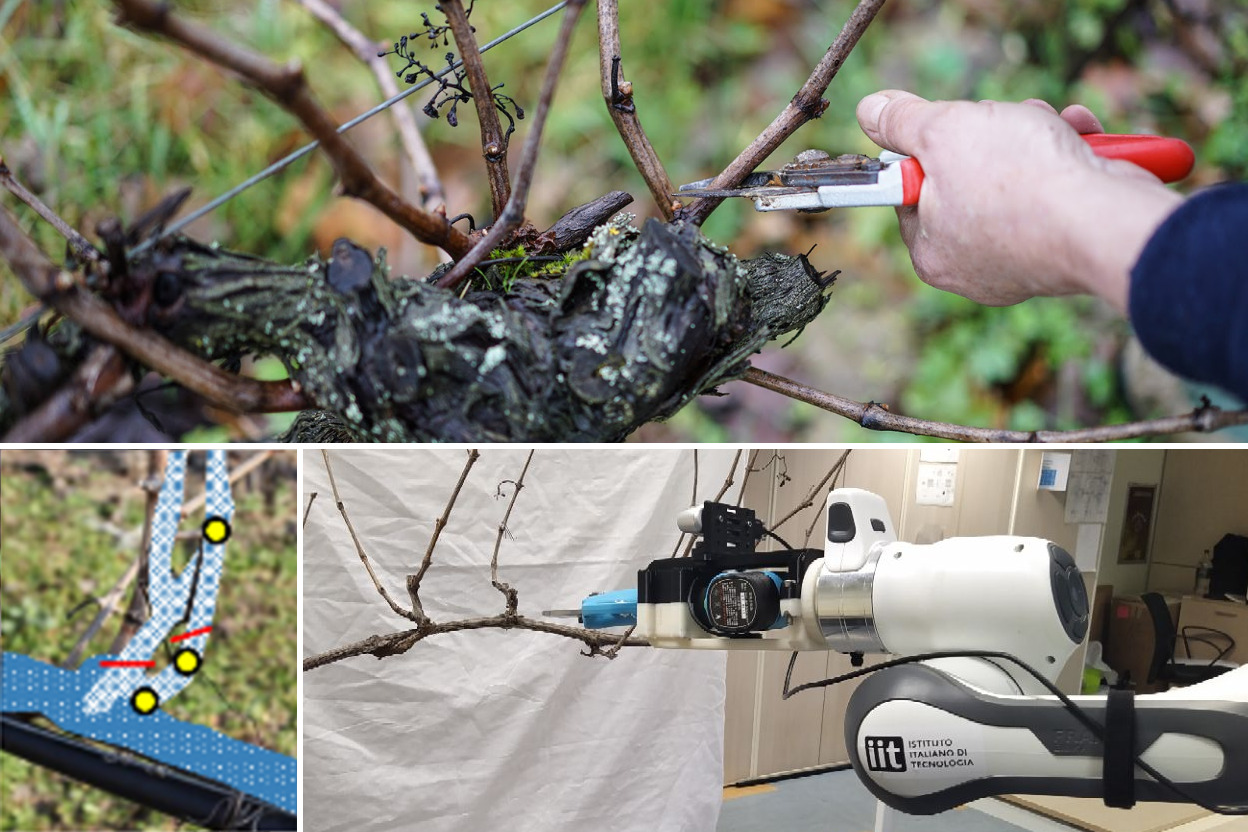}
    \caption{The top half shows an example of manual grapevine pruning and the bottom half presents our robot performing grapevine pruning along with the desired plant modeling, with the dark blue representing a main cordon, light blue the canes, yellow the nodes and red mark the pruning points.}
    \label{fig:intro:pruning}
\end{figure}

Automation with agri-food robots to accomplish various tasks in the field is a long-time challenge recognized by the community. A recent review article about agricultural robotics \cite{Vougioukas2019} states that perception is a significant challenge in the field.
Robot perception is the ability for the robot to understand the environment and the objects (crops, fruits, etc) that it has to deal with. The robot visual perception in our application refers to object segmentation, finding the object boundaries in an image, and object detection, the recognition of the interested parts of the objects inside a given image.
An important task to perform in a vineyard is winter pruning, a complex operation that needs to be completed during the dormant season \cite{PONI201688}. Performing a balanced winter pruning allows a good compromise between remunerative yield and desired grape quality hence maximizing grower's income \cite{Intrieri116, PONI2018445}.
Figure \ref{fig:intro:pruning} shows a comparison between a human-performed pruning and a robot-performed pruning, showing similarities.

\begin{figure*}[t]
    \centering
    \includegraphics[width=\textwidth]{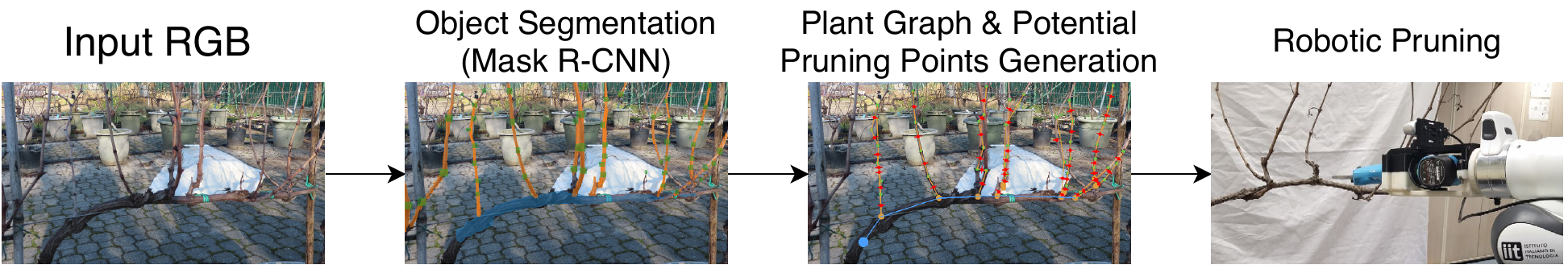}
    \caption{Robotic grapevine winter pruning flowchart. The meaning of the colors in the 2\textsuperscript{nd} and 3\textsuperscript{rd} step are explained later in Fig. \ref{fig:dataset:annotation_exmpl} and Fig. \ref{fig:dataset:output_exmpl}}
    \label{fig:intro:pipeline}
    \vspace{-20pt}
\end{figure*}

To perform this task in an autonomous way, there are some steps that need to be considered, such as creating an object segmentation dataset where the primary object is the grapevine in a dormant season scenario. This dataset will require more time and effort for annotation with respect to annotating the bounding boxes only.
The second step is the training of a neural network for object segmentation, where the data that are available may not be accurate enough, due to large variations in vine age and size, training and the inherent randomness of nature.
After the neural network is trained with this dataset, a third step is the usage of its output to generate the potential pruning points. This can be done by analyzing the network inference, connecting the different grapevine organs to each other in a graph based structure that allows the generation of the desired potential pruning points on a spur-pruned grapevine.
Figure \ref{fig:intro:pipeline} shows the entire pipeline used by our approach.

The main contribution of this paper is the use of a representative model describing the actual grapevine architecture.
This 2D model allows the identification of potential pruning points by using the plant structure information encoded in it, without the need to use feature descriptors (for points of interest detection), feature matching algorithms or stereo camera systems (for 3D reconstruction).

As such, we propose the creation of an object segmentation dataset, described in Section \ref{sec:dataset}, a neural network fine tuning to perform object segmentation is described in Section \ref{sec:neuralnet}, along with the plant graph generation and potential pruning points detection in Section \ref{sec:grapgen}.
We later demonstrate our experimental setup, discussing the achieved results in Section \ref{sec:results}.
In the end, we conclude showing advantages and disadvantages, giving an overall summary, of our approach and presenting possible future work in Section \ref{sec:conclusions}.

\section{Related Work}
In the deep learning field, various object segmentation\cite{Nogues2018}, detection\cite{Zhang2018}, and tracking algorithms\cite{Santos2019} have been introduced by the computer vision and robotics communities in the past years. Some of these studies have also been applied on the agri-food field, such as fruit detection \cite{Borianne2019} for yield estimation purposes \cite{Bargoti2016}, weed removal \cite{DiCicco2016, Milioto2017}, plant phenotyping \cite{Grimm2018} dealt with potential pruning points generation in long-cane pruned grapevines \cite{doi:10.1002/rob.21680}.
The authors in \cite{DiCicco2016} deal with the challenge introduced by the variety of work regions in the fields, lighting, weather conditions, which leads to the difficulties for semantic segmentation of crop fields.
The authors in \cite{Grimm2018} present a proof of concept for detecting and quantifying plant organs for yield estimation without using destructive means. This approach is based on automated detection, localization, count and analysis of plant parts used to estimate yield.
The authors in \cite{Santos2019} present a new public dataset with grape clusters annotated in 300 images and a new annotation with interactive image segmentation to generate object masks, a new public dataset for grape detection and instance segmentation containing images, bounding boxes, masks and an evaluation of two state-of-the-art methods for object detection, object segmentation and a fruit counting methodology. 
Authors in \cite{doi:10.1002/rob.21680} used a trinocular stereo cameras system and correspondence algorithms to obtain a 3D reconstruction of the plant, useful to compute the pruning points in the wild.

These representative works demonstrated several important concepts, such as real-time plant segmentation \cite{Milioto2017}, the more in-depth plant feature extraction for finding multiple parts of the plant\cite{Grimm2018}, and an annotation tool to generate a segmentation dataset quickly\cite{Santos2019} and a pruning generation system.
However, an approach to prune without the use of complex stereo camera systems or the usage of mobile platform that covers the grapevine, is still missing in the literature.
To the best knowledge of the authors, this is the first time that a successful robot is able to prune a grapevine using a mobile platform carrying only a robotic arm, with a pruning tool and a depth camera.



\section{Dataset}
\label{sec:dataset}
First of all, a dataset is needed to train the neural network for grapevine segmentation. We decided to create a dataset with three classes, the main cordon, the cane and the node.
An example of these annotation concepts can be seen in Fig. \ref{fig:dataset:annotation_exmpl}. The main cordon is the horizontal static structure and the canes are the plant organs that normally show a vertical orientation. The nodes are structures present on the canes where new shoots may grow. Dividing the grapevine into these three main categories allows us to generate potential pruning points.

\subsection{Data Acquisition}
\label{sec:datatset:aquisition}
The data acquisition was performed in an experimental vineyard of one of the Vinum project partner.
Data was captured using a common 4k Canon compact camera, by recording a video closeup of a grapevine row segment. The frames from the video were extracted using \texttt{ffmpeg} into a total of 171 frames. These frames have a resolution of 3840$\times$2160 pixels. Currently we are only using these images, with the possibility in the future to add more via additional capture or by performing data augmentation.


\subsection{Data Annotation}
\label{sec:dataset:annotation}
\begin{figure}[tpb]
    \centering
    \includegraphics[width=0.42\textwidth]{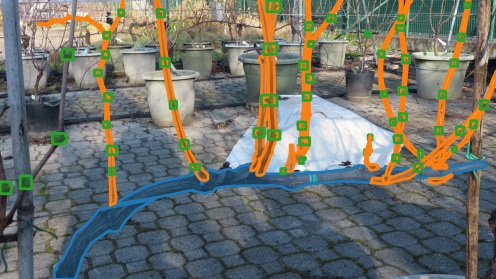}
    \caption{Example of an annotation, where blue represents the main cordon class, orange the cane class and green the node class.}
    \label{fig:dataset:annotation_exmpl}
\end{figure}

The dataset is being annotated following the COCO segmentation dataset, with the three mentioned classes, the main cordon, the cane and the node.
We chose the COCO format for the annotation since it is a common format for segmentation annotation, which contributes to a higher availability of annotation tools that use this format, as well as some neural network frameworks provide built-in processing of this format.
The annotation tool being used is COCO annotator, a web-based tool that is designed for efficiently label images. This tool also includes the possibility of connecting to an external neural network allowing automatic annotation of an image using a pre-trained neural network, discussed in Section \ref{sec:neuralnet}. This feature has been used for aiding in the annotation of more samples, as described in Section \ref{sec:neuralnet}.
The 171 captured images were annotated and then split into 136 training image and 35 evaluation images. We are not using a test set due to the fact that the actual test is going to be preformed with the actual robot on a vineyard.
The images are being annotated in their original size, allowing to downscale the image as required. As mentioned, Fig. \ref{fig:dataset:annotation_exmpl} presents an example of the annotations created, where blue represents the main cordon, orange represents the canes and green represents the nodes.
These images are being annotated by two different persons, although none of them being an expert in the area, and in the future the dataset shall be reviewed and annotated by experts in the area.

\section{Neural Network}
\label{sec:neuralnet}
In order to solve the proposed task, we used a deep neural network that is able to perform object detection and segmentation.
The currently used framework is Detectron2\cite{wu2019detectron2}, 
Facebook AI Research's implementation of state-of-the-art detection algorithms, using Pytorch and including a model zoo with baselines trained, presenting the obtained result metrics. These metrics are bounding box and mask Average Precision,  train time per iteration, inference time per image and the required memory for the training.
The base network that is being used is the Mask R-CNN with several backbones models. The first being tested was R50-FPN, a Residual Network with 50 layers, combined with Feature Pyramid Networks. 
For experimentation two additional models were tested, the R101-FPN and the X101-FPN, where the first is a Residual Network with 101 layers with the same structure as the R50-FPN and the second is an ResNeXt with 101 layers. 
This ResNext is an improvement on the original ResNet network, considering a new dimension named cardinality on top for the normal height and depth of a neural network, which is the size of the set of transformations.

The network is being trained using the default training procedure of Detectron2. This procedure creates a model, optimizer, scheduler and dataloader with the default configurations provided along with the model. It then loads the pre-trained model weights, initializes logging functions and starts to follow a standard training workflow with a single-optimizer single-datasource iterative optimization. The training hyperparameters are the default ones, with the only changes being the batch size changed to 4, from original value 16, and the number of training iterations that was changed to 50000, from the original 270000. These changes were due to time and hardware constraints.


\begin{figure}[tpb]
    \centering
    \includegraphics[width=0.42\textwidth]{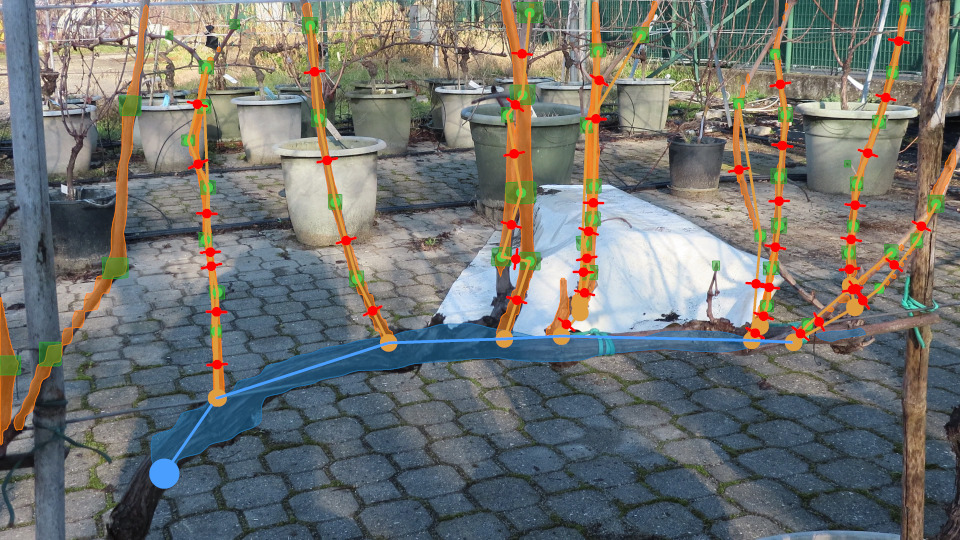}
    \caption{An output example of potential pruning points generation overlaid on the segmentation output, where blue represents the main cordon class, orange the cane class and green the node class, with the red markers representing the generated potential pruning points.}
    \label{fig:dataset:output_exmpl}
\end{figure}

\section{Plant Graph Generation and Potential Pruning Points Detection}
\label{sec:grapgen}

After obtaining the inference produced by the trained neural network, an additional layer of processing is needed to find the desired potential pruning points. The method that we decided to use was the creation of a processing layer that interprets this inference by understanding how the several segments are related to each other.

\subsection{Plant Graph Generation}
The data structure used to host these data is a tree-shaped graph, with the main cordon as the root element. Then, there are canes connected to the main cordon, or to other canes. The leaf elements are nodes, or canes that have no nodes on them.
The important concept that we want to analyze is the relation between canes, and nodes on the canes, in order to obtain accurate potential pruning points. Figure \ref{fig:dataset:output_exmpl} presents an example of a created plant graph overlaid on the segmentation output.
This tree-shaped graph connects the various grapevine inferred items, showing the topographical structure of the plant. Each grapevine item consists of a unique identifier number, bounding box coordinates, score, segmentation mask, class identifier, class name, class color, item color, center, thickness, distance from the parent, depth and parent.
This item is based on the common aspects of the neural network classes, and extended depending on the class, where the main cordon contains a list of canes, sorted by their distance to the parent, the canes have a list of nodes, sorted using the same metric. In the end, the graph's root node is the main cordon item.

There are three sets of connections in the graph structure, the ``main cordon to canes'' set, relating the main cordon to its connected canes, the ``cane-to-cane proximity'' set relating each cane to its proximal canes and the ``cane to nodes'' set relating each cane to its connected nodes.

\begin{algorithm}[t]
    \SetAlgoLined
    \Parameter{$dilation, max\_iter, n$}
    \Input{$masks_A \rightarrow{\text{Mask group A}}$}
    \Input{$masks_B \rightarrow{\text{Mask group B}}$}
    \Output{$connections \rightarrow{\text{Map of connections}}$}
    \caption{Get connections between two sets of masks}
    \ForEach{$mask \in masks_B$} {
        \While{$iter \leq max\_it \textbf{ and not}(connected)$} {
            \If{$iter > 0$} {
                Dilate $mask$
            }
            Get the indices of $mask$'s non-zero values\\
            Obtain the corresponding values on $masks_A$\\
            \If{There are non zero correspondences} {
                Get the lowest correspondence\\
                Get the intersection with $mask$\\
                \If{Intersection is in the $n^{th}$ slot} {
                    Add connection to $connections$\\
                    Set connected to true\\
                }
            }
            $iter++$
        }
    }
    \label{alg:MainCordonToShoot}
\end{algorithm}

Algorithm \ref{alg:MainCordonToShoot} presents a generic method used to associate the canes to their respective main cordon. It achieves this by using an auxiliary matrix containing all main cordon instances, with shape $n_{MC} \times H \times W$, where $n_{MC}$ is the number of main cordons inferences, $H$ and $W$ being the height and width of the input image. As \ref{eq:1} shows, each main cordon segmentation mask ($mask$) is inserted into the matrix ($masks_A$) with the corresponding identifier number ($ID_{MC}$), increased by 1, leaving 0 as the background class, at its non-zero values.

\begin{equation}
\begin{gathered}
    \forall mask \in masks_{MC}: \\ masks_A[i] = mask * (IDs_{MC}[i] + 1), \\ i=1, ..., n_{MC} 
\end{gathered}
\label{eq:1}
\end{equation}
For each cane segmentation mask the row and column indices of its non-zero values are searched and then used to get the unique values in the matrix at these indices, across all the main cordon segmentation masks. This way we can know which is the overlapping main cordon. Then, dividing the cane mask into $n$ different slots, arranged in a vertical manner, we select the main cordon that overlaps the cane in the $n^{th}$ slot, if it is present. It may happen that the cane does not overlap with any main cordon inference due to the imperfection of the inference, and as such, a solution was found performing an incremental dilation of the cane segmentation mask until a main cordon is found or a maximum number of dilations is reached. The number of dilations, the size of the dilation and the number of vertical slots are user-defined parameters for this algorithm.

The same method can be iteratively used to find the connections between the proximal canes.
The initial search field is composed only by the group of canes already connected to the main cordon. At each iteration new connections are searched for the non connected canes among the connected canes (the search field), using the same overlapping and dilation concepts.
These newly found connections are then added to the final connections set and the two groups of masks are updated.
The iteration stops when no new connections are found or all the canes have been connected.
The algorithm starts by considering the output of Algorithm \ref{alg:MainCordonToShoot} as $masks_A$ and the set difference between all cane masks and $masks_A$ as $masks_B$. It also takes as input the size of the dilation ($dilation$), the maximum number of dilations ($max\_it$) and the number of vertical slots ($n$) and it outputs a map whose keys are cane identifiers and values are lists of cane identifiers.

The last set of connections is obtained using a variation of Algorithm \ref{alg:MainCordonToShoot}, where, instead of checking if the intersection is contained in a specific part of the mask, the most overlapping cane is considered as connected cane.
The algorithm takes as input the list of cane masks ($masks_A$) and the list of node bounding boxes ($masks_B$) and it outputs a map whose keys are cane identifiers and values are lists of node identifiers.

\subsection{Potential Pruning Points Detection and Localization}
With the previously structure created, the generation of pruning points can be performed.
Currently, we decided to use a crude approach for detecting potential pruning points, which are points on canes, either between two nodes of the same cane, between the bases of two canes growing from the same cane, between the base of a cane and its first node. An example of this can be seen in Fig. \ref{fig:dataset:output_exmpl}, indicated by the red markers.
As \ref{eq:2} shows, by default, a potential pruning point ($\vec{pp}$) is the midpoint between two of the previously mentioned points ($\vec{p_1}$ and $\vec{p_2}$).
For selecting the final pruning point, during this initial work, we decided to select the pruning point located above the second node of a cane.

Due to the possible curvature of the canes, it may happen that this midpoint is not contained in the cane mask, and if this happens, the point is moved to a point inside the mask.
In the end, an orientation angle is needed, to orient the pruning tool roll angle, in order to perform the cut correctly.
This orientation angle ($\alpha$) is computed taking into account the slope angle between the straight line connecting the two points and the horizontal direction of the image.

\begin{equation}
    \begin{gathered}
        \forall (\vec{p_1}, \vec{p_2}) \\
        \vec{pp} = \frac{\vec{p_1} + \vec{p_2}}{2} \\
        \delta_{x, y} = {p_1}_{x, y} - {p_2}_{x, y} \\
        \alpha = \begin{cases}
        0, & \mbox{ if } \delta_x = 0 \\
        \frac{\pi}{2}, & \mbox{ if } \delta_y = 0 \\
        \arctan{\frac{\delta_y}{\delta_x}} - \mathop{\mathrm{sign}}{\frac{\delta_y}{\delta_x}} * \frac{\pi}{2}, & \mbox{otherwise}
        \end{cases}
    \end{gathered}
    \label{eq:2}
\end{equation}

\section{Experimental Setup and Results}
\label{sec:results}
In this section, we describe how the experiments have been carried out, showing first two different testing environments and then illustrating the actual achieved performances of our approach.

\subsection{Experimental Setup}
\begin{figure}[t]
    \begin{subfigure}{0.23\textwidth}
        \includegraphics[height=6.6cm]{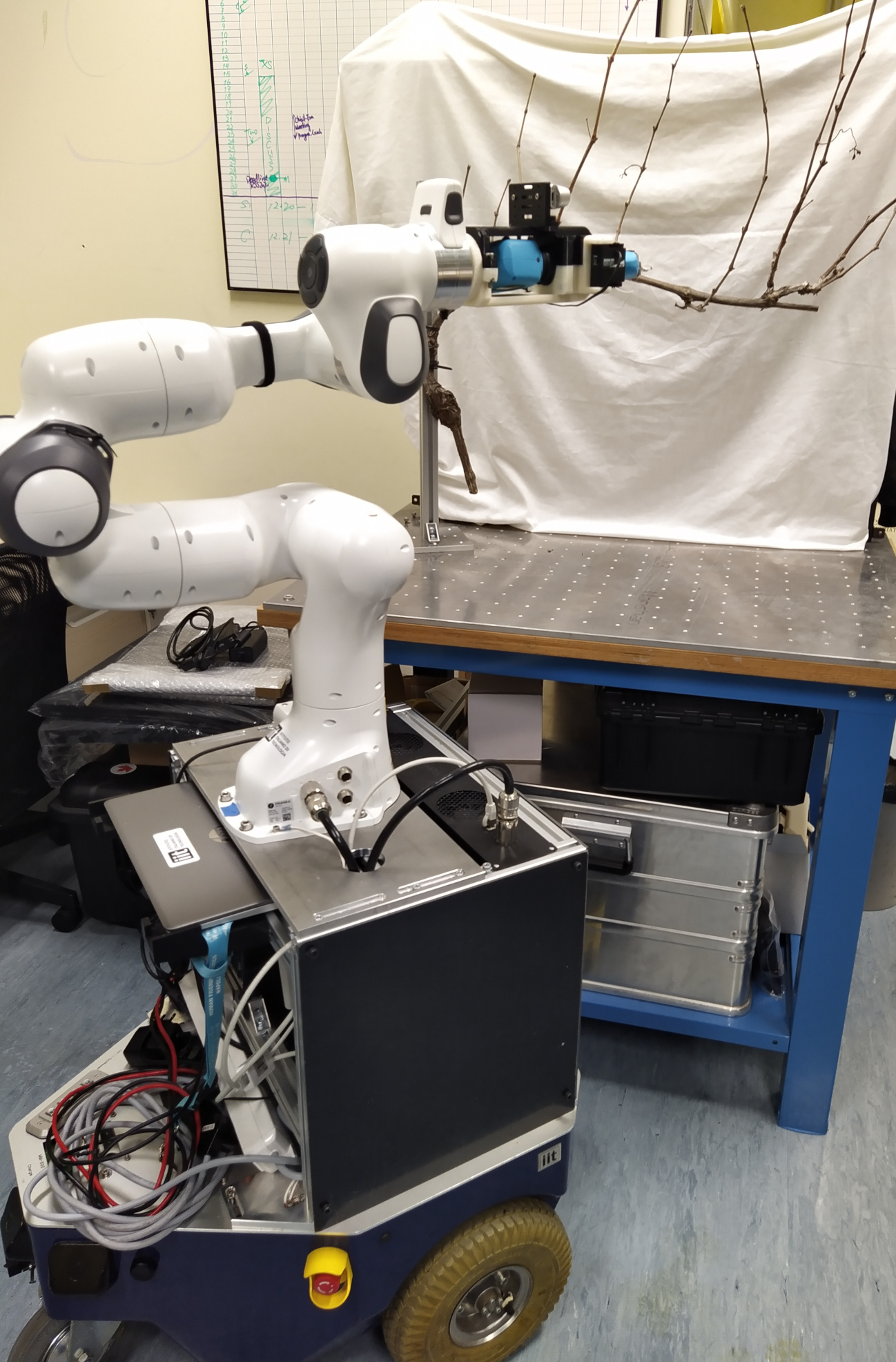}
    \end{subfigure}
    \begin{subfigure}{0.23\textwidth}
        
        \includegraphics[width=\textwidth]{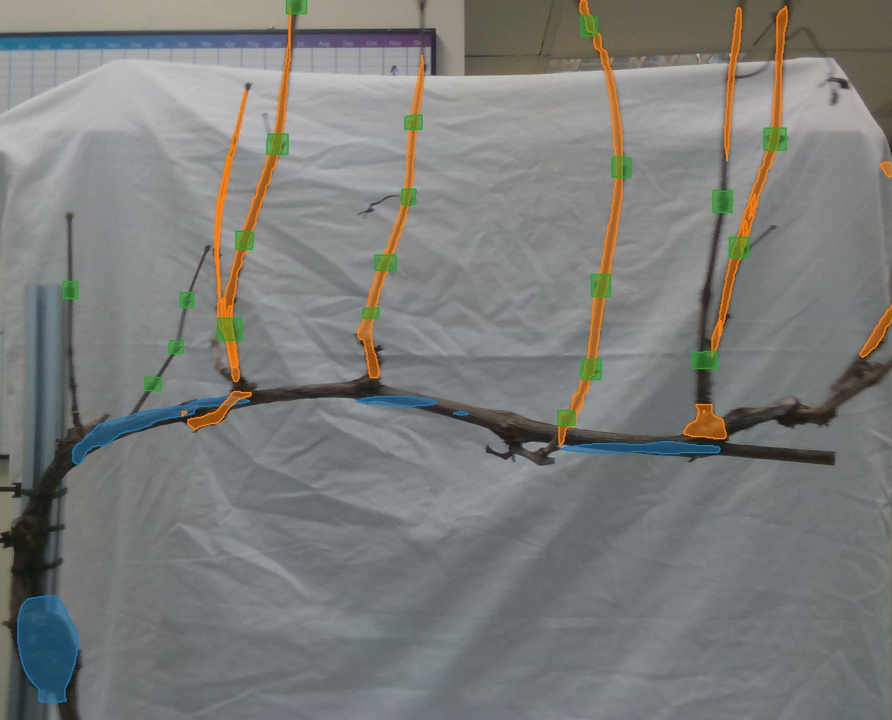}
        \includegraphics[width=\textwidth]{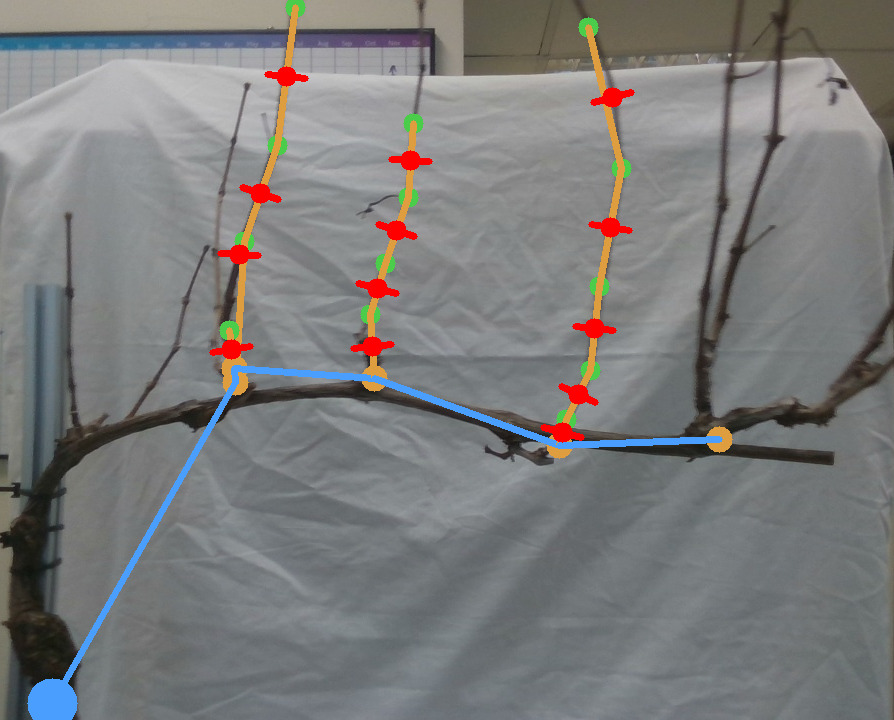}
        
    \end{subfigure}
    \caption{Experimental setup containing one of the grapevine specimens and the target robot.}
    \label{fig:intro:rlsetup}
\end{figure}

The experimental setup for testing the segmentation neural network is split into two parts. 
The first part is a testing setup that uses images captured in the fields. These images were captured in a natural environment without changes that would modify the environment, the same kind of environment a farmer would work on.
The inference is performed by the neural network on these images, and uses the graph generation algorithm presented in the previous section. These images do not interact with the neural network training, being only used for visual output evaluation. The second experimental setup was created using real grapevines in a lab environment, using a grapevine specimen that allows the emulation of the grapevine environment in a laboratory environment. This allows the testing of the complete potential pruning points detection pipeline in a safe manner, as shown in Fig. \ref{fig:intro:rlsetup}, allowing an additional layer of testing, by evaluating the performance of both backgrounds.

\begin{figure}[t]
    \centering 
    \includegraphics[width=0.42\textwidth]{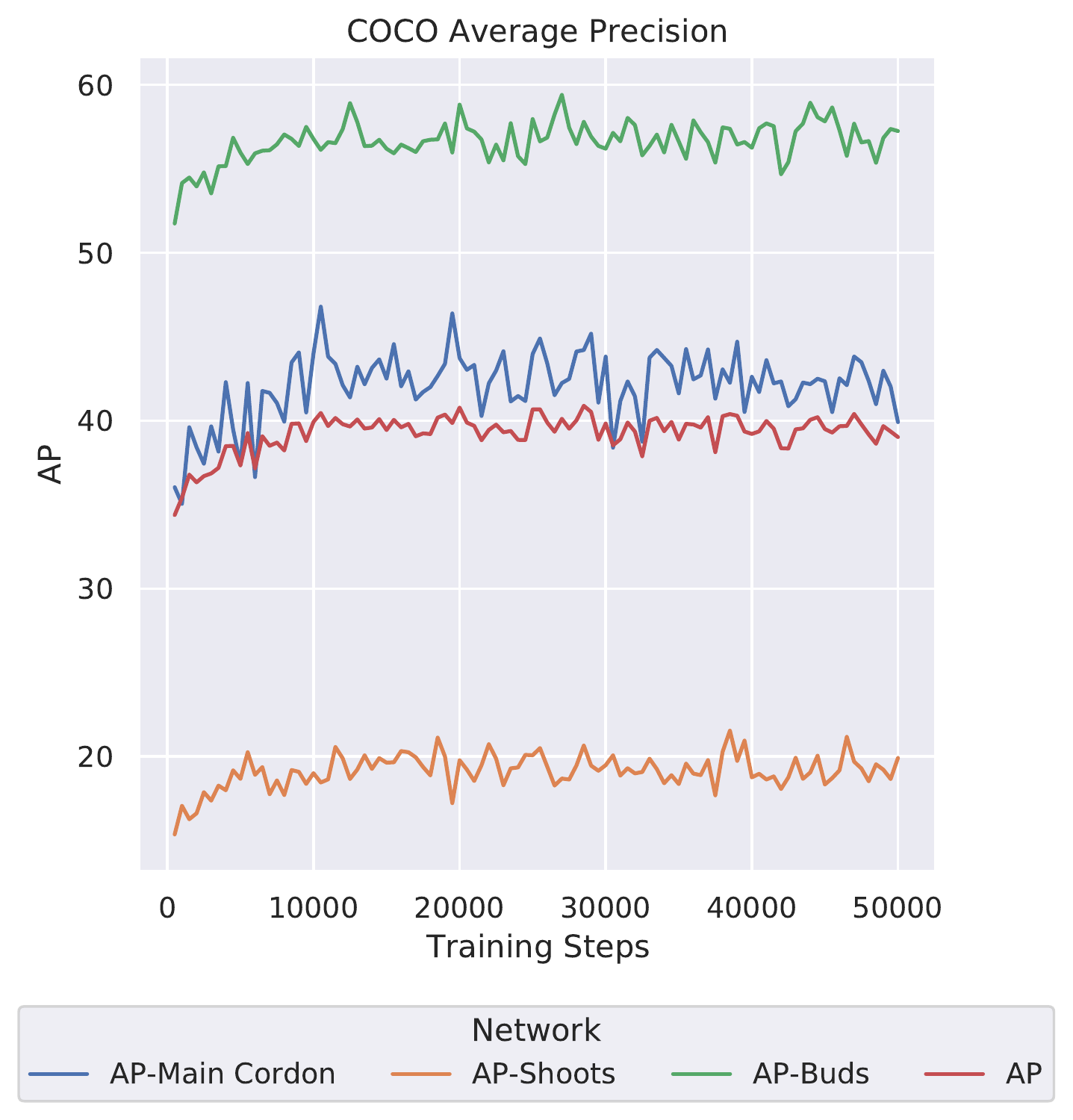}
    \caption{The validation results for the network model Resnet 50 along the training process, following COCO evaluation method, presenting the Average Precision (AP) of each singular class and the mean Average Precision of all classes.}
    \label{fig:results:nntr50classes}
\end{figure}

\begin{table}[t]
    \caption{Average Precision and Average Recall after training for the three tested models. Both metrics are calculated with intersection over union IoU=0.50:0.90, considering all area sizes and a maximum detections value of 100.}
    \label{tab:results:nntv}
    \begin{center}
        \begin{tabular}{|c|c|c|c|}
            \hline
            & Resnet 50 & Reset 101 & ResNeXt 101\\
            \hline
            Average Precision & 39.0\% & 41.1\% & 41.6\% \\
            \hline
            Average Recall & 47.0\% & 48.7\% & 48.2\% \\
            \hline
        \end{tabular}
    \end{center}
\end{table}

\subsection{Experimental Results}

\begin{figure*}[t]
    \centering
    
    \includegraphics[width=0.185\textwidth]{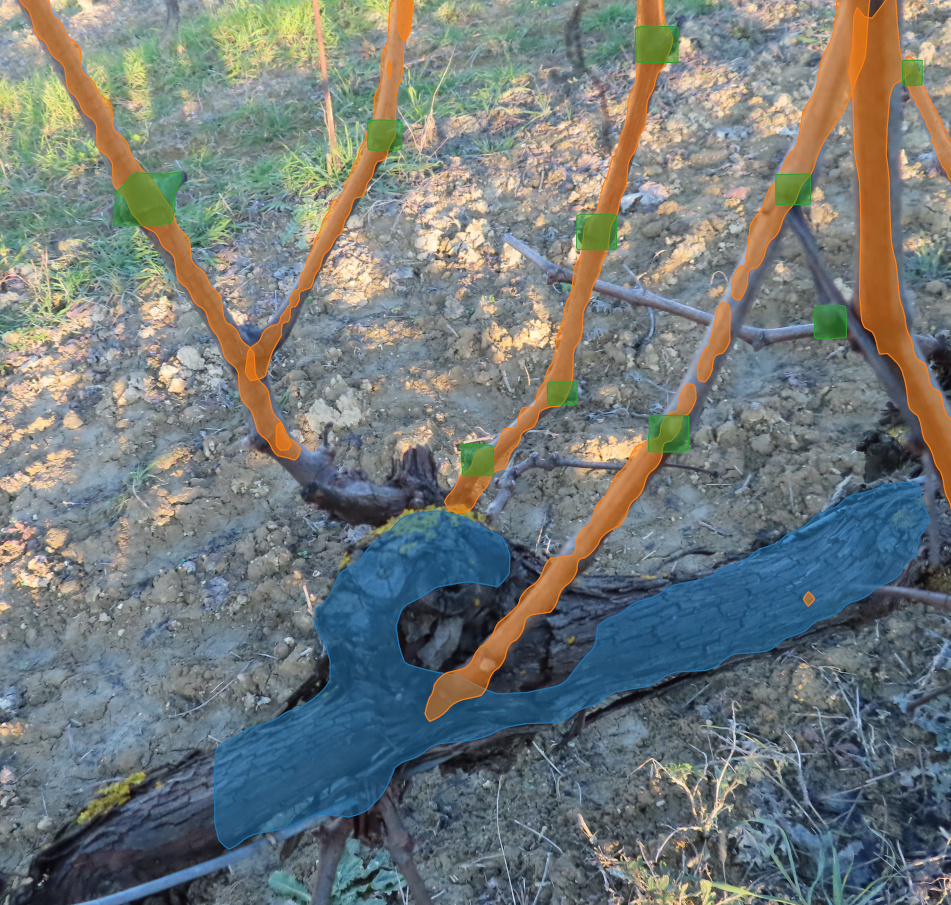}
    \includegraphics[width=0.185\textwidth]{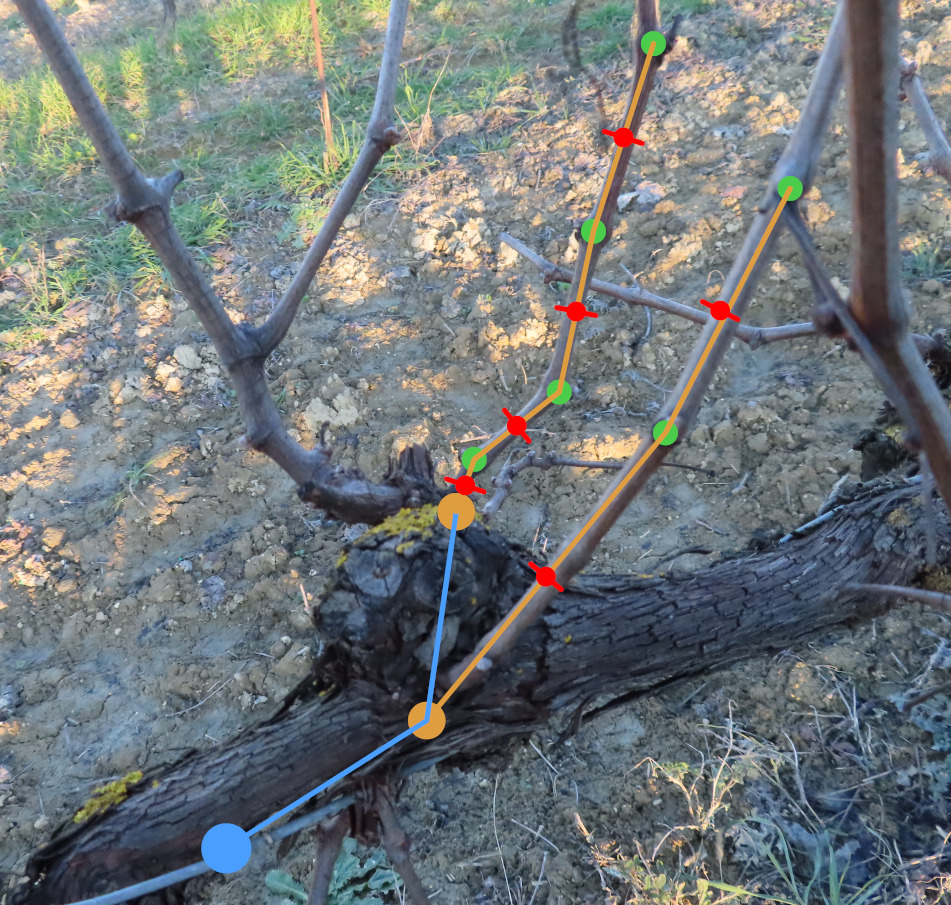}
    \includegraphics[width=0.305\textwidth]{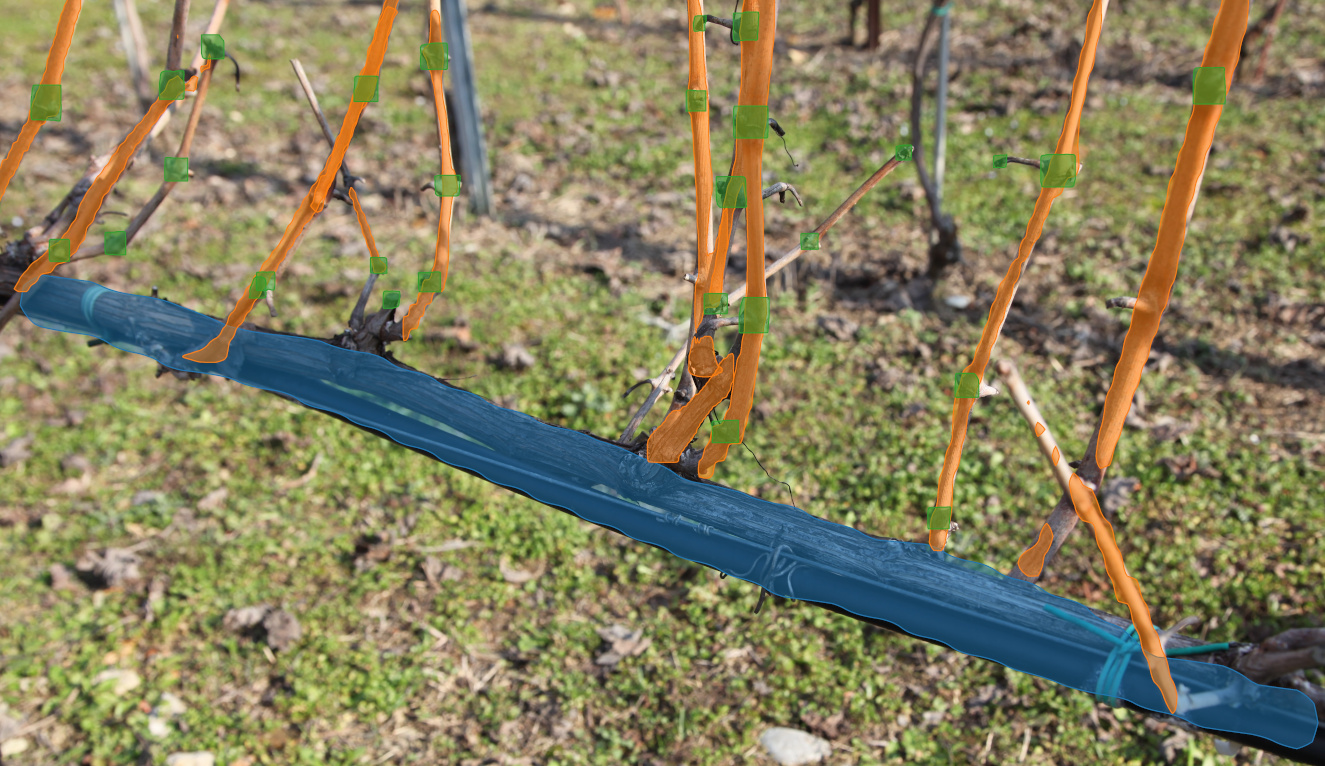}
    \includegraphics[width=0.305\textwidth]{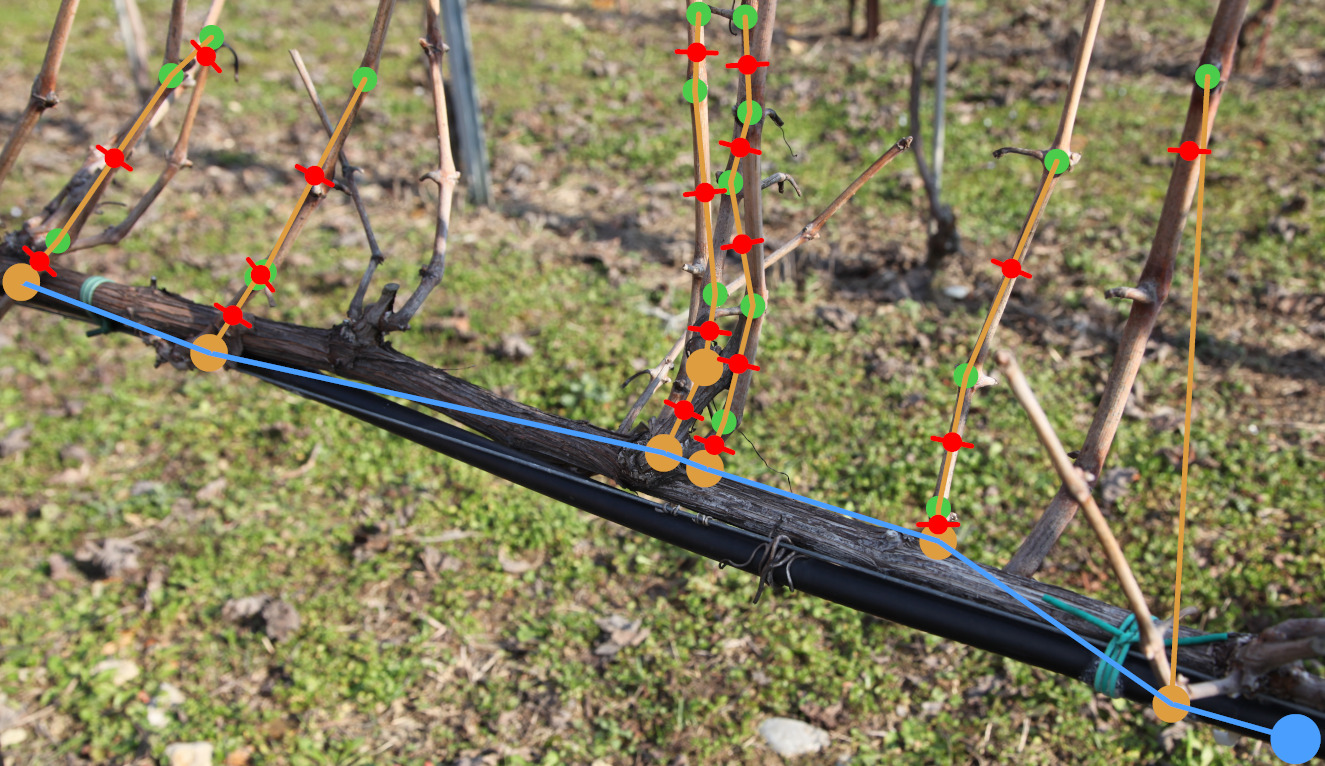}

    \caption{Two pairs of images, where the first image of the pair shows the inference output and the second image shows plant graph and potential pruning points generation. As in the previous figures, the color blue represents the main cordon class, the color orange the various detected canes, the green the nodes. The red markers represent the generated potential pruning points. }
    \label{fig:results:cap_img}
    \vspace{-20pt}
\end{figure*}

This section is split into two parts, one to show the neural network training trend and its  validation results and the second one to give a qualitative evaluation about the plant graph generation algorithm and the subsequent potential pruning points generation.

\subsubsection{Network Training and Validation}

Figure \ref{fig:results:nntr50classes} shows the various Average precision (AP) metrics calculated during the training process. The AP metric presented in the figure is on par with the Detectron2 COCO instance segmentation baseline, where the segmentation AP is around 40\% in our training where the baseline value is 37.2\%.  It is also noticeable the difference between the various classes, which can be explained by the difficulty of each class, starting from the easiest, the nodes are closer to bounding boxes than segmentation, since the node is a specific structure along the cane. Due to the nature of the training process performed on the grapevine by the grower, the main cordon is a well-defined structure present on the grapevine, being mostly on the same place, with lesser variation of its shape. For the canes, it is an object that is harder to perceive due to being more prone to occlusion by other canes, blending with the background due to its thinness and its randomness. Table \ref{tab:results:nntv} show the comparative results of the three tested models after training, presenting both Average Precision and Average Recall.

\subsubsection{Plant Graph Creation and Potential Pruning Points Generation}

The potential pruning points detection performance is related to the graph generation performance, which is dependent on the segmentation network performance.
This is due to each connection being related to a potential pruning point, except for the connections between main cordons and canes.
In particular, each missed connection leads to a missed potential pruning point, i.e. a false negative sample.
Non-existent connections may be established, leading to extra-detected pruning points, i.e. false positive samples.
In the case of cane connections, the actual connection rule is based on checking if the masks are intersected in their bottom part, since, as said, canes tend to grow upwards.
With this, canes that grow downwards, although rare, are not detected.

\subsection{Results Visual Evaluation}

When considering the performance of the several created components, it is also important to understand how it performs on the field. An example of this can be seen in Fig. \ref{fig:results:cap_img}, where although it does not find every single cane, the graph generation still manages to find viable potential pruning points that can be used.
It is important to note that the images were not used to train the network, neither were acquired with the same camera as the one used on the robot. These results may not represent the complete challenge the neural network using images captured by the robot may face, but are still important to analyze.
Considering our main goal to develop automated winter spur pruning of grapevines, an important thing to understand is how the neural network performs using the camera equipped on the robot. An example of this is shown in Fig.\ref{fig:intro:rlsetup}.
An important thing to notice is the maturity difference between grapevine specimens from the acquired data and the grapevine specimen present in the lab, where the one in the lab is a much younger plant, leading to a thinner main cordon and canes, which affects the segmentation of the grapevine, affecting mostly the main cordon. Nevertheless, the system still is able to generate valid potential pruning points.

\section{Discussion and Conclusions}
\label{sec:conclusions}

Due to the imperfect nature of the obtained inference, the detected grapevine items, created and used for potential pruning points generation, may not be accurate. This imperfection is caused by multiple factors, such as the lack of data on grapevine specimens with various ages or the capture conditions. To mitigate this, data augmentation may be considered in order to expand the existing dataset, as well as capturing new images from different grapevines.
However, even considering imperfect segmentation, we are able to create a 2D plant's structure model. The plant model, and consequently the pruning points, is heavily dependent on the segmentation, which could lead to the non detection of potential pruning points.
Furthermore, the algorithm that establishes the connections can be seen as a crude method, that does not solve certain anomalies that may happen, leading to false connections being created. However, this initial solution still allows the robot to autonomously prune the grapevine.

In conclusion, in this paper we presented a novel method to create 2D plant models, based on grapevines semantic segmentation, containing the topographical and geometrical information between the different grapevine organs.
We demonstrated how our approach is able to create a significant set of potential pruning points. The final set of the real pruning points can be selected. In this initial phase, the selection consists on the pruning point located above the second node of a cane.

Future work will revolve on the improvement of the inference results, by continuing to explore the data augmentation, by adjusting the functions used and the order they are applied.
On the graph generation side, the connection algorithm can be replaced by a CNN that takes as input two masks (a two-channel black and white image) and outputs a connection confidence measure, which can be used to decide whether to establish the connection between the items or not.
Moreover, improvements on the logic for pruning points detection can be carried out, by adding information such as canes thickness, for vigor evaluation, and improving items localization, for a better assessment of the growth direction.








\bibliographystyle{IEEEtran}

\bibliography{root}

\begin{thebibliography}{10}
\providecommand{\url}[1]{#1}
\csname url@samestyle\endcsname
\providecommand{\newblock}{\relax}
\providecommand{\bibinfo}[2]{#2}
\providecommand{\BIBentrySTDinterwordspacing}{\spaceskip=0pt\relax}
\providecommand{\BIBentryALTinterwordstretchfactor}{4}
\providecommand{\BIBentryALTinterwordspacing}{\spaceskip=\fontdimen2\font plus
\BIBentryALTinterwordstretchfactor\fontdimen3\font minus
  \fontdimen4\font\relax}
\providecommand{\BIBforeignlanguage}[2]{{%
\expandafter\ifx\csname l@#1\endcsname\relax
\typeout{** WARNING: IEEEtran.bst: No hyphenation pattern has been}%
\typeout{** loaded for the language `#1'. Using the pattern for}%
\typeout{** the default language instead.}%
\else
\language=\csname l@#1\endcsname
\fi
#2}}
\providecommand{\BIBdecl}{\relax}
\BIBdecl

\bibitem{Vougioukas2019}
\BIBentryALTinterwordspacing
S.~G. Vougioukas, ``Agricultural robotics,'' \emph{Annual Review of Control,
  Robotics, and Autonomous Systems}, vol.~2, no.~1, pp. 365--392, 2019.
  [Online]. Available:
  \url{https://doi.org/10.1146/annurev-control-053018-023617}
\BIBentrySTDinterwordspacing

\bibitem{PONI201688}
\BIBentryALTinterwordspacing
S.~Poni, S.~Tombesi, A.~Palliotti, V.~Ughini, and M.~Gatti, ``Mechanical winter
  pruning of grapevine: Physiological bases and applications,'' \emph{Scientia
  Horticulturae}, vol. 204, pp. 88 -- 98, 2016. [Online]. Available:
  \url{http://www.sciencedirect.com/science/article/pii/S0304423816301650}
\BIBentrySTDinterwordspacing

\bibitem{Intrieri116}
\BIBentryALTinterwordspacing
C.~Intrieri and S.~Poni, ``Integrated evolution of trellis training systems and
  machines to improve grape quality and vintage quality of mechanized italian
  vineyards,'' \emph{American Journal of Enology and Viticulture}, vol.~46,
  no.~1, pp. 116--127, 1995. [Online]. Available:
  \url{https://www.ajevonline.org/content/46/1/116}
\BIBentrySTDinterwordspacing

\bibitem{PONI2018445}
\BIBentryALTinterwordspacing
S.~Poni, M.~Gatti, A.~Palliotti, Z.~Dai, E.~Duchêne, T.-T. Truong, G.~Ferrara,
  A.~M.~S. Matarrese, A.~Gallotta, A.~Bellincontro, F.~Mencarelli, and
  S.~Tombesi, ``Grapevine quality: A multiple choice issue,'' \emph{Scientia
  Horticulturae}, vol. 234, pp. 445 -- 462, 2018. [Online]. Available:
  \url{http://www.sciencedirect.com/science/article/pii/S0304423817307586}
\BIBentrySTDinterwordspacing

\bibitem{Nogues2018}
\BIBentryALTinterwordspacing
F.~C. Nogues, A.~Huie, and S.~Dasgupta, ``{Object Detection using Domain
  Randomization and Generative Adversarial Refinement of Synthetic Images},''
  may 2018. [Online]. Available: \url{http://arxiv.org/abs/1805.11778}
\BIBentrySTDinterwordspacing

\bibitem{Zhang2018}
\BIBentryALTinterwordspacing
Y.~Zhang, K.~Jia, and Z.~Wang, ``{Part-Aware Fine-grained Object Categorization
  using Weakly Supervised Part Detection Network},'' \emph{IEEE Transactions on
  Multimedia}, pp. 1--1, jun 2019. [Online]. Available:
  \url{http://arxiv.org/abs/1806.06198
  http://dx.doi.org/10.1109/TMM.2019.2939747}
\BIBentrySTDinterwordspacing

\bibitem{Santos2019}
\BIBentryALTinterwordspacing
T.~T. Santos, L.~L. de~Souza, A.~A. dos Santos, and S.~Avila, ``{Grape
  detection, segmentation, and tracking using deep neural networks and
  three-dimensional association},'' \emph{Computers and Electronics in
  Agriculture}, vol. 170, jul 2020. [Online]. Available:
  \url{http://arxiv.org/abs/1907.11819
  http://dx.doi.org/10.1016/J.COMPAG.2020.105247}
\BIBentrySTDinterwordspacing

\bibitem{Borianne2019}
\BIBentryALTinterwordspacing
P.~Borianne, F.~Borne, J.~Sarron, and E.~Faye, ``{Deep Mangoes: from fruit
  detection to cultivar identification in colour images of mango trees},'' sep
  2019. [Online]. Available: \url{http://arxiv.org/abs/1909.10939}
\BIBentrySTDinterwordspacing

\bibitem{Bargoti2016}
\BIBentryALTinterwordspacing
S.~Bargoti and J.~P. Underwood, ``Image segmentation for fruit detection and
  yield estimation in apple orchards,'' \emph{Journal of Field Robotics},
  vol.~34, no.~6, pp. 1039--1060, 2017. [Online]. Available:
  \url{https://onlinelibrary.wiley.com/doi/abs/10.1002/rob.21699}
\BIBentrySTDinterwordspacing

\bibitem{DiCicco2016}
M.~{Di Cicco}, C.~{Potena}, G.~{Grisetti}, and A.~{Pretto}, ``Automatic model
  based dataset generation for fast and accurate crop and weeds detection,'' in
  \emph{2017 IEEE/RSJ International Conference on Intelligent Robots and
  Systems (IROS)}, 2017, pp. 5188--5195.

\bibitem{Milioto2017}
A.~{Milioto}, P.~{Lottes}, and C.~{Stachniss}, ``Real-time semantic
  segmentation of crop and weed for precision agriculture robots leveraging
  background knowledge in cnns,'' in \emph{2018 IEEE International Conference
  on Robotics and Automation (ICRA)}, 2018, pp. 2229--2235.

\bibitem{Grimm2018}
\BIBentryALTinterwordspacing
J.~Grimm, K.~Herzog, F.~Rist, A.~Kicherer, R.~T{\"{o}}pfer, and V.~Steinhage,
  ``{An Adaptive Approach for Automated Grapevine Phenotyping using VGG-based
  Convolutional Neural Networks},'' nov 2018. [Online]. Available:
  \url{http://arxiv.org/abs/1811.09561}
\BIBentrySTDinterwordspacing

\bibitem{doi:10.1002/rob.21680}
\BIBentryALTinterwordspacing
T.~Botterill, S.~Paulin, R.~Green, S.~Williams, J.~Lin, V.~Saxton, S.~Mills,
  X.~Chen, and S.~Corbett-Davies, ``A robot system for pruning grape vines,''
  \emph{Journal of Field Robotics}, vol.~34, no.~6, pp. 1100--1122, 2017.
  [Online]. Available:
  \url{https://onlinelibrary.wiley.com/doi/abs/10.1002/rob.21680}
\BIBentrySTDinterwordspacing

\bibitem{wu2019detectron2}
Y.~Wu, A.~Kirillov, F.~Massa, W.-Y. Lo, and R.~Girshick, ``Detectron2,''
  \url{https://github.com/facebookresearch/detectron2}, 2019.

\end{thebibliography}

\end{document}